\documentclass{article}
\usepackage{graphicx} 
\usepackage{tikz}
\usetikzlibrary{trees}
\usepackage{amsmath,amssymb}
\usepackage{amsfonts}
\usepackage{bbm}
\usepackage[square,numbers]{natbib}
\usepackage{authblk}
\usepackage{adjustbox}
\usepackage{tcolorbox}
\usepackage{caption} 
\usepackage{titlesec}

\graphicspath{{figs/}}

\titleformat{\section}
  {\normalfont\large\bfseries} 
  {\thesection}                
  {1em}                        
  {}                           

\titleformat{\subsection}
  {\normalfont\normalsize\bfseries} 
  {\thesubsection}
  {1em}
  {}

\title{The Role of Predictive Uncertainty and Diversity \\in Embodied AI and Robot Learning}
\author{Ransalu Senanayake}
\affil[]{\small Arizona State University\footnote{Email: ransalu@asu.edu, ransalu@cs.stanford.edu} }
\date{}

\begin{document}

\maketitle

\begin{abstract} 

Uncertainty has long been a critical area of study in robotics, particularly when robots are equipped with analytical models. As we move towards the widespread use of deep neural networks in robots, which have demonstrated remarkable performance in research settings, understanding the nuances of uncertainty becomes crucial for their real-world deployment. This guide offers an overview of the importance of uncertainty and provides methods to quantify and evaluate it from an applications perspective.

\end{abstract}

\section{\large Why Do We Need (or Not Need) Uncertainty?}

\begin{figure}[]
\includegraphics[width=1.0\columnwidth]{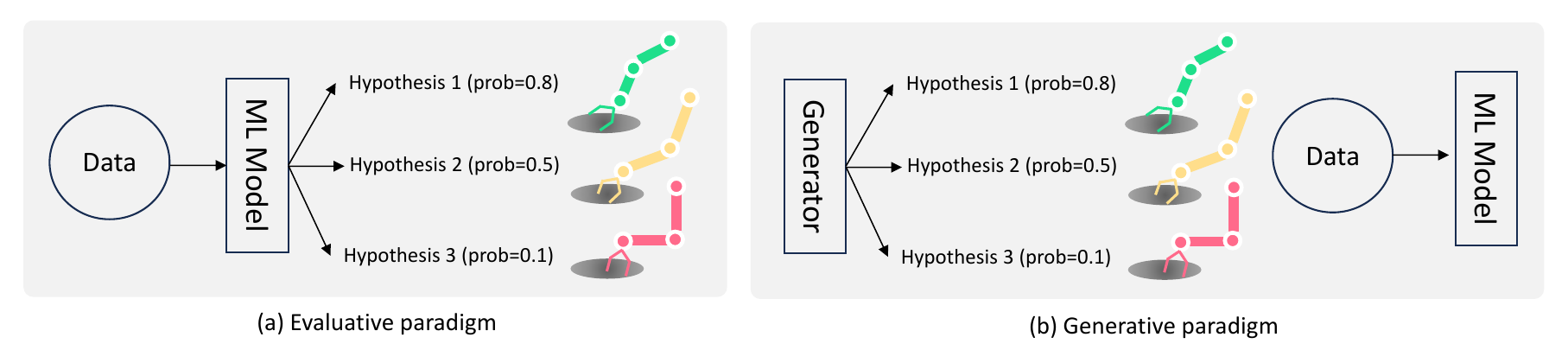}
\caption{Two paradigms of diversity. (a) In the \emph{evaluative paradigm}, the machine learning model provides various hypothesis with associated likelihoods based on data it has seen. (b) In the \emph{generative paradigm}, we need to generate hypothetical outcomes.}
\label{fig:diver}
\end{figure}

We want our robots to \emph{efficiently learn} how to \emph{robustly act} in the unrestrained physical world. A robot's ability to experience and think through \emph{diverse} possible scenarios---whether in perception or action---and their consequences helps enhance robustness and generalizability. In principle, countless conceivable scenarios can exist. However, learning all possible scenarios or gaining a complete understanding of them is not only impossible but often redundant. Instead, empowering the robots with the ability to discern what is important and what is not facilitates efficient learning and scrutiny of their own decisions. Having a sense of the \emph{likelihood} of these scenarios aids in prioritizing their importance.

Characterizing the likelihood of events and actions about robots and the physical world they operate in inherently involves uncertainty, whether represented as probabilities, sets, or any other form. However, some techniques for handling uncertainty tend to be computationally expensive and sometimes even inaccurate, thereby defeating the purpose of working with uncertainty for efficient learning and enhancing robustness. This issue becomes particularly pronounced as models grow larger, especially with limitations of hardware in embodied agents. Therefore, in any given application, it is crucial to balance the trade-off among accuracy, uncertainty, and computational complexity. 

\begin{figure}[]
\centering
\includegraphics[width=\columnwidth]{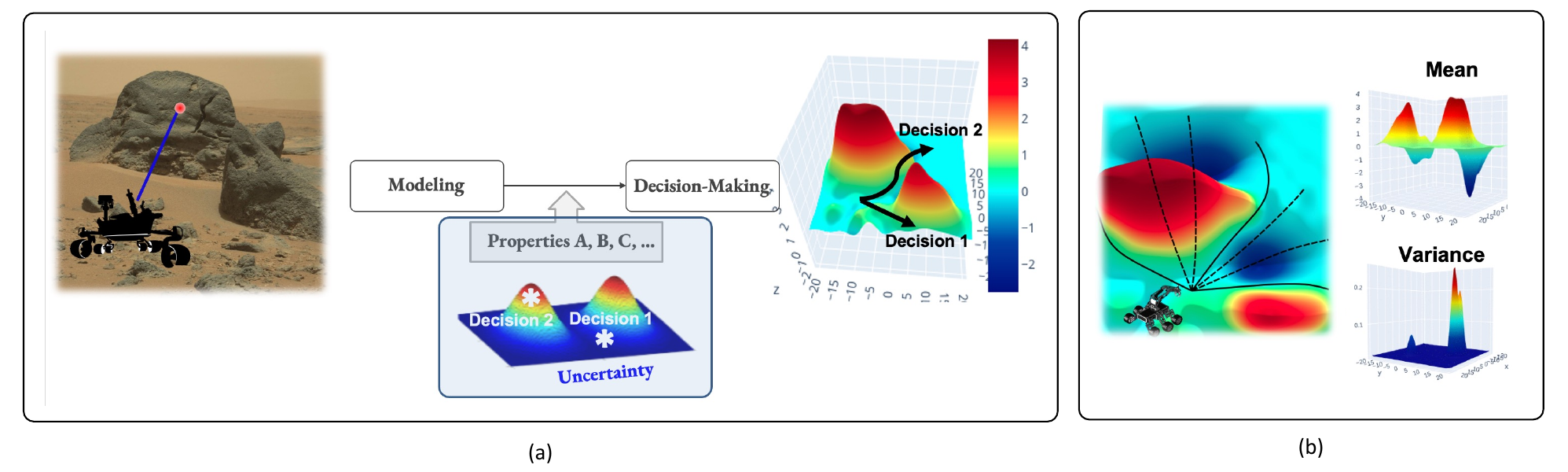}
\caption{An example of epistemic uncertainty. (a) A rover builds an elevation map of Mars~\cite{martian} to make decisions. If we have a distribution of maps instead of a single map, we can have different decision options, for instance to minimize risk, power consumption, etc. (b) The pits behind the boulders are not visible from the rover's front view. However, if the rover quantifies the \emph{epistemic uncertainty}, then it knows how to make safe decisions. To represent this uncertainty, in its simplest form, we need the mean and variance of elevation~\cite{senanayake2017bayesian}.}
\label{fig:rover_dec}
\end{figure}

As shown in Figure~\ref{fig:diver}, there are at least two main paradigms where the diversity and uncertainty of predictions prove beneficial: \emph{evaluative paradigm} and \emph{generative paradigm}. In the evaluative paradigm, the robot learns a model that captures its uncertainty about the world. For instance, consider the mapping example illustrated in Figure~\ref{fig:rover_dec}. If the robot cannot observe a particular area of the environment, say, due to occlusion, the model should quantify the uncertainty of that area as high. With this uncertainty, we can generate multiple maps with varying degrees of likelihood. In probabilistic terms, if we have a map that represents mean and variance, finitely many maps can be sampled from that map representation. Given the geography, a 10 $km$ deep pit behind the boulder is unlikely. However, the presence of flat ground or even a 2 $m$ deep pit behind the boulder is more plausible. Thus, access to uncertainty enables the generation of many such hypotheses. Having access to such diverse hypotheses aids in devising various decision options, which can then be evaluated to make the optimal decision. When the robot is capable of quantifying the uncertainty of the world, it can make risk-averse or risk-seeking decisions~\cite{nishimura2023rap}, which in turn helps with certifying robots and building trustworthiness.

In the generative paradigm, our objective is to generate diverse worlds, scenarios, or data. Generation can be performed using a machine learning (ML) model, such as text2video~\cite{videoworldsimulators2024}, or a digit twin, such as a physics-based simulator~\cite{li2021igibson, szot2021habitat} or virtual reality~\cite{suomalainen2020virtual}. The generated cases can either be used to learn a machine learning model~\cite{kurenkov2020visuomotor,wang2023voyager} or to test an existing model~\cite{Delecki2022arxiv,Sagar2024Failures}. The former application is particularly popular in robotics, as collecting data from simulators is cost-effective and efficient~\cite{makoviychuk2021isaac}. Further, simulation is also invaluable in scenarios where factors such as safety~\cite{mun2023user,li2024choose} and algorithmic fairness~\cite{Pathiraja2024Fairness} is paramount, for example, in autonomous driving~\cite{Pathiraja2024Fairness,dosovitskiy2017carla}. The diversity and uncertainty of simulation also facilitates the learning of robust ML models, for example, through domain randomization, as demonstrated in frameworks such as BayesSim~\cite{ramos2019bayessim}. As the other application of the generative paradigm, we can generate edge cases for identifying failure modes~\cite{Delecki2022arxiv} and out-of-distribution scenarios~\cite{nitsch2021out,mcallister2019robustness,farid2022task}.

\section{When Do We Need (or Not Need) Uncertainty?}

We already discussed that uncertainty helps with making more informed decisions, learning robust models, and testing existing pre-trained models. Uncertainty, also known as ambiguity, stems from the stochasticity of the world. Stochasticity, also known as randomness, is a comfort term we often use when we do not know how to model the reality\footnote{This statement is partially biased towards the philosophy of \emph{determinism} though there are many arguments that supports the philosophy of \emph{indeterminism}.} or we do not care about perfectly modeling the reality. Attempting to deal with all sources of stochasticity is often wasteful in real-time applications such as robotics. Therefore, it is important to understand the potential sources of uncertainty in embodied AI agents.



\subsection{The Sources of Uncertainty}
\label{sec:sources}

\begin{figure}[]
\includegraphics[width=1.0\columnwidth]{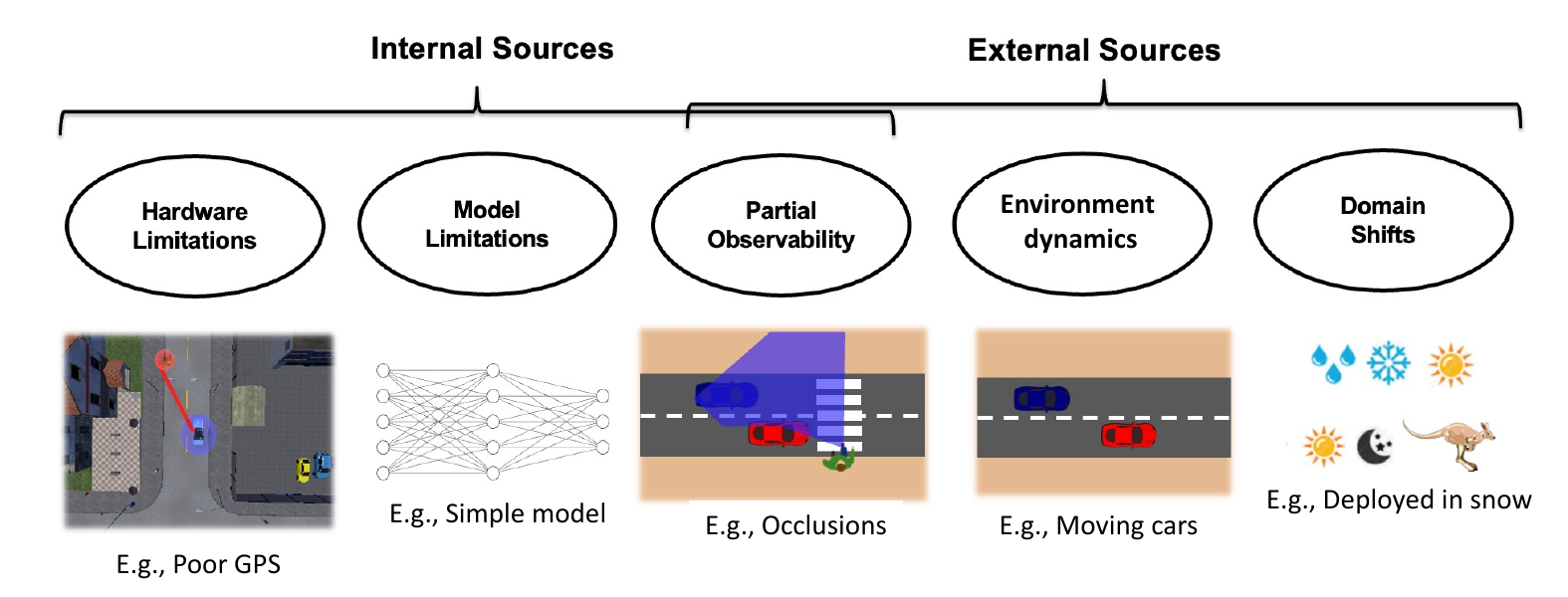}
\caption{Sources of uncertainty.}
\label{fig:sources}
\end{figure}

As also illustrated in Figure~\ref{fig:sources}, broadly speaking, uncertainty could be due to internal sources or external sources:
\begin{enumerate}
    \item Physical limitations: Uncertainty can arise from errors in measurement devices, actuators, and human inputs. All measurement devices exhibit systematic, random, and gross errors. These errors can compound, exacerbating the uncertainty. For example, consider an autonomous vehicle that attempts to localize a pedestrian using a pre-computed map. The vehicle's LIDAR system introduces measurement error when determining the distance to the pedestrian. Additionally, the vehicle's precise location may be uncertain due to poor GPS signals around high-rise buildings, further increasing the uncertainty of the pedestrian's location estimation in the map. As another example, robot actuators, such as stepper motors, may not execute commands with perfect accuracy. This inaccuracy can worsen due to mechanical wear and tear, operation outside specified conditions (such as overloading or extreme temperatures), and calibration issues. Uncertainty can also stem from the way humans abstract their knowledge through models. When humans provide prior knowledge to models, in terms of preferences~\cite{ouyang2022training,biyik2023active} or Bayesian priors~\cite{gelman1995bayesian}, this information may be incomplete and prone to errors.
    \item Model limitations: Machine learning models are increasing in size, yet the constraints of limited onboard hardware and the need for real-time inference necessitate the use of smaller models. However, opting for simpler models or quantizing larger models to reduce their size invariably leads to errors and high uncertainty~\cite{mayoral2023robotperf}. 
    \item Partial observability: Occlusion~\cite{senanayake2017bayesian}, environmental factors such as fog or ambient darkness~\cite{Pathiraja2024Fairness}, and clutter~\cite{laskey2016robot} can introduce uncertainty into a robot's observations. Employing multiple sensors or sensor modalities can sometimes mitigate this uncertainty~\cite{manyika1995data}. Factors such as human intentions may also be considered partially observable~\cite{xu2019learning,luo2018porca}. Partial observations can arise from hardware limitations or a model's inability to process them, representing more internal sources of uncertainty. Conversely, changes in the operating environment can be viewed as an external source of uncertainty. Partial observability is explicitly addressed in certain decision-making models, such as Partially Observable Markov Decision Processes (POMDPs)~\cite{kochenderfer2015decision,kaelbling1998planning}.
    \item Environment dynamics: The environment in which a robot operates changes over space and time. For instance, objects on a tabletop might move and people in a house or street may walk. Such changes invariably result in uncertainty~\cite{itkina2019dynamic,vintr2019spatio,senanayake2017learning}.
    \item Domain shifts: When a robot encounters situations not covered during the training of its neural networks, the robot becomes more uncertain. For instance, if an autonomous vehicle is trained exclusively with data from Arizona, it may struggle to operate in Boston, where snow is common~\cite{Delecki2022arxiv,nitsch2021out,Pathiraja2024Fairness}, or in Australia, where it might encounter unfamiliar animals such as kangaroos~\cite{Deahl2017Volvo}.
\end{enumerate}

\subsection{A Few Examples of Uncertainty in Embodied AI}

Let us now discuss a few examples of how these various sources of uncertainty affect three robotics applications.\\

\noindent \emph{Vision-language-based navigation}: With the advancement of multi-modal ML models, the development of mobile robots capable of interacting with and assisting humans is becoming increasingly feasible. Currently, preliminary versions of such robots are already prevalent, serving purposes such as cleaning and providing aid to humans, exemplified by devices such as Amazon Astro and Roomba connected to Alexa. Consider a robot designed to navigate within homes. In order to navigate, it first needs to construct a map of the house. Traditionally, this involves creating an occupancy map, a process that requires exploring uncertain (i.e., unseen) areas of the house. In a house, walls are typically fixed in place, furniture is semi-permanent, and humans and pets are the most dynamic elements. Robots may encounter challenges such as becoming stuck, performing unsafe actions, or facing tasks that are beyond their physical capabilities---for instance, a Roomba cannot open doors, whereas a Spot robot can. Moreover, these robots will need to incorporate large language models (LLMs) and vision-language models (VLMs) to interact with human to understand their unknown intents and behaviors. Further, these large models needs to be small enough, by construction or quantization, to fit in limited hardware. However, reducing the model size to accommodate computing constraints compromises model performance and increase in uncertainty of estimates~\cite{team2023gemini}.\\

\noindent \emph{Manipulation}: Many robotics tasks in the future will require manipulation. Robots need to plan how to move the end effector of a robot arm from place A to place B, and then grasp it. Uncertainty in motion planning could be due to unknown or occluded objects along the robot's path. When grasping an object, objects can be occluded or the object shape can be uncertain~\cite{li2016dexterous}. Further, the deformability, type of material, unknown mass and friction, all can contribute to uncertainty. Even if these factors are known, the pose can be uncertain~\cite{stulp2011learning, hsiao2011robust}, especially for mobile manipulators such as a manipulator on a quadruped. Multiple sensor modalities such as vision and touch helps to reduce the uncertainty. Uncertainty can also arise when the object is dynamic, for instance when picked up an object from a conveyor belt or from another robot or human~\cite{bowman2019intent}.\\

\noindent \emph{Field robotics}: Field robots are primarily designed for outdoor operations. They take various forms such as drones, copters, rovers, boats, legged robots with arms, autonomous underwater vehicles (AUVs), and other specialized configurations. These robots find applications in a range of activities, including surveillance, environmental monitoring, mining, agriculture, and exploration in underwater or space environments. The unstructured nature of these environments introduces significant uncertainty into their operation. Further, in environments like mines, underwater, or space, GPS signals are unavailable, necessitating the estimation of the robot's location using its trajectory or nearby landmarks through techniques such as Simultaneous Localization and Mapping (SLAM)~\cite{durrant2006simultaneous}. However, SLAM does not typically account for uncertainties caused by occlusions and domain shifts. As another source of uncertainty, dynamic elements such as moving people and objects can make field robotics tasks challenging, especially in urban environments. Control also becomes particularly difficult in conditions where control commands may not be executed as expected due to air or water turbulence, or slippery surfaces caused by snow.\\

\section{How Do We Quantify Uncertainty?}

In embodied agents, we obtain data and then fit a ML model. Irrespective of the sources discussed in Section~\ref{sec:sources} and how well the model is trained, no model is perfect. Therefore, we need to quantify the uncertainty of the predictions. However, how we can to quantify this uncertainty depends on the type of uncertainty.

\subsection{Types of Uncertainties: The Known Unknowns and Unknown Unknowns}

\begin{figure}[h]
\includegraphics[width=1.0\columnwidth]{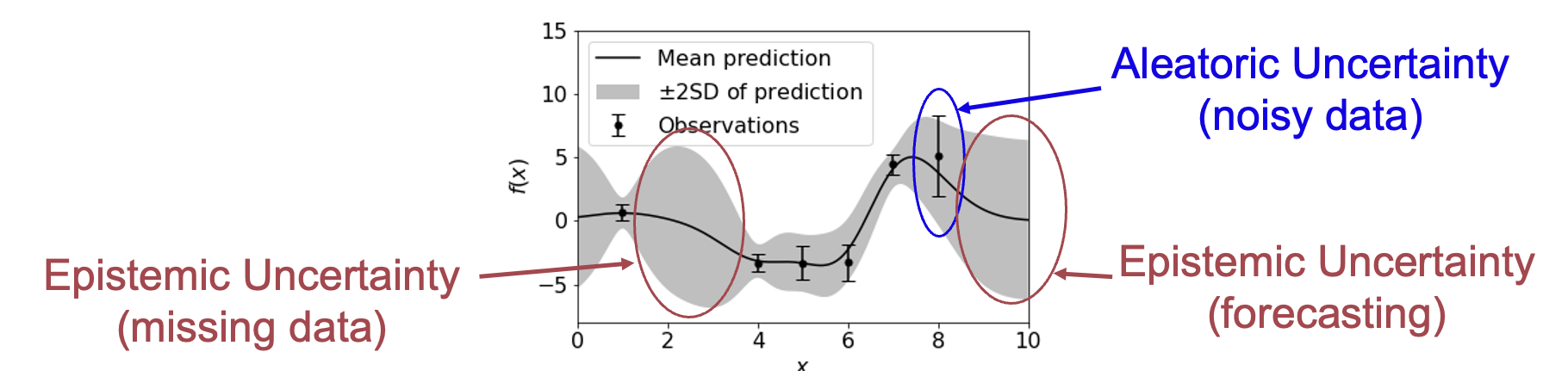}
\caption{Types of uncertainty. Aleatoric uncertainty (known unknowns) is due to the inherent randomness of data whereas epistemic uncertainty (unknown unknowns) is due to lack of data. If we do not have data in a particular region of the input space or if we try to predict the future, the epistemic uncertainty is high.}
\label{fig:plot_unc}
\end{figure}

Uncertainty can be categorized into two types: 
\begin{enumerate}
    \item Aleatoric uncertainty (known unknowns) - This uncertainty is also known as statistical uncertainty. It represents the inherent randomness of a system and cannot be reduced with more data. Aleatoric uncertainty is widely studies in probabilistic robotics~\cite{thrun2002probabilistic} and is less challenging to work with. 
    \item Epistemic uncertainty (unknown unknowns) - This uncertainty is also known as model uncertainty or systematic uncertainty. Since it stems from the lack of knowledge about the world, the more data we collect, the more we can reduce this uncertainty. Epistemic uncertainty is not straightforward to estimate. 
\end{enumerate}

To explain the distinction between the two types of uncertainties more pictorially, consider Figure~\ref{fig:plot_unc}. When we take measurements for a given \(x\), the measurements vary slightly each time. This randomness, depicted by error bars in the plot, is known as aleatoric uncertainty as it is the inherent noise. In regions where we lack data, $(x,y)$, whether we are interpolating or extrapolating, the epistemic uncertainty is high. The further we are from data, the higher the uncertainty becomes. Estimating aleatoric uncertainty is relatively straightforward, and most frequentist statistical methods address it. However, quantifying epistemic uncertainty involves representing multiple models through a set, an ensemble, or a probability distribution. In deep ensembles~\cite{lakshminarayanan2017simple} and Monte Carlo dropout, it is obvious that there are multiple models because they consider different weight combinations of the neural network. When a probability distribution is introduced over weights of a neural network, as in Bayesian methods, it implicitly creates an infinite number of neural networks where some models are more likely than others. 

The term \emph{Bayesian} is typically an overloaded term in robotics. In some classical application areas of robotics such as filtering, the term Bayesian is used whenever the Bayes' theorem is applied. In more learning tasks, a distribution over the parameters of the ML model is introduced and the Bayes' theorem is used to estimate the parameter distribution given data. In some sub-communities of Bayesian statistics, the term Bayesian is used only when an approximate Bayesian inference techniques such as Markov chain Monte Carlo or variational inference is used to solve a complex problem with many priors and hyperpriors. Under such nomenclature, even vanilla Gaussian processes (GPs), which are considered as a Bayesian nonparamatric technique in statistical ML, are not Bayesian enough as they do not introduce parameters over the hyperparameters of the GP kernels.

\subsection{Measuring Uncertainty: Metrics}
\label{sec:metrics}

Historically, most uncertainty estimates are represented as probabilities though other forms exist. From a \emph{measure theory} perspective, a probability measure $\mu$ is a real-valued function defined in a $\sigma$-algebra, while satisfying Kolmogrov axioms of 1) non-negativity, 2) unit measure, and 3) countable additivity~\cite{berger2001statistical}. In other words, a probability measure assigns a real number in the range $[0,1]$ to each event in the $\sigma$-algebra, such that the measure of the entire sample space is 1. These probabilities can be used to develop metrics to measure uncertainty in different ways.\\

\noindent {\bf Variance}: It measures the \emph{uncertainty of a random variable}. A high variance indicates high uncertainty as it represents the dispersion (i.e., how far) from the mean. For a discrete random variable $X$ with outcomes $\{x_1, x_2, ..., x_N\}$, associated probabilities $\{p_1, p_2, ..., p_N\}$, and mean $\mu = \sum_{n=1}^N p_n x_n$, the variance can be computed as,
\begin{equation}
    \sigma^2 = \sum_{n=1}^N p_n (x_n - \mu)^2.
\end{equation}
\\

\noindent {\bf Entropy}: It measures the \emph{uncertainty of the possible outcomes of a random variable or a system}. As the entropy represents the expected amount of information that is needed to describe the random variable $X$, it can be computed as,
\begin{equation}
    H(X) = -\sum_{n=1}^N p_n \log p_n.
\end{equation}
Entropy only considers the probability of outcomes rather than outcomes themselves. Entropy is measured in bits for the logarithmic base of 2. The more uncertain it is, the higher the entropy is. In other words, if every outcome is equally probable, then the entropy is high. For instance, for a binary classifier with the probability of the positive class $p$, the entropy can be computed as $-( p\log p + (1-p) \log (1-p) )$. This entropy is $0$ when the probability is $1$ or $0$ (i.e., very much certain) while it achieves its highest when the probability is 0.5 (i.e., most uncertain).\\

\noindent {\bf Negative Log Probability}: The negative log-likelihood (NLL) measures the \emph{correctness of a classifier's prediction compared to the ground truth}. For a softmax output $p$ of the ground truth class, the negative log-likelihood (NLL)\footnote{In classification settings with softmax outputs, the equation of NLL is similar to that of cross entropy.} can be computed as $-\log p$. If a ML model outputs high probability to the correct class, then the NLL value will be closer to 0. Conversely, if it outputs a low probability to the correct class, the NLL value will be high. Therefore, the lower the NLL, the better the model is. 

The prediction of a ML model can be a probability distribution rather than a scalar. In such cases, NLL can be used to measures the \emph{correctness of a classifier's predictive distribution compared to the ground truth}. If the predictive distribution of a ML model is a Gaussian with mean $y_*$ and standard deviation $\sigma_*$, the negative log probability~\cite{seeger2004gaussian} can be computed as,
\begin{equation}
    -\log p(y_* | y_t) = \frac{1}{2} \log (2\pi \sigma_*^2) + \frac{(y_* - y_t)^2}{2 \sigma_*^2}
\end{equation}
for ground truth value $y_t$.\\

\noindent {\bf Mahalanobis Distance}: The Mahalabonis distance measures the \emph{distance between a point and a probability distribution}. It is defined as,
\begin{equation}
    \sqrt{(\mathbf{x}-\mu)^\top \Sigma^{-1} (\mathbf{x}-\mu)}
\end{equation}
for a point $\mathbf{x} \in \mathbb{R}^d$ from a multivariate distribution with mean $\mu  \in \mathbb{R}^d$ and covariance matrix $\Sigma^2 \in \mathbb{R}^{d \times d}$. When $d=1$, the distance is $||x - \mu||_2 / \sigma$. In regression, the Mahalnobis distance can be used for measuring the distance between the ground truth value and prediction represented by the mean and variance. It is also used in out-of-distribution detection~\cite{nitsch2021out}.\\

\begin{figure}
\centering
\caption*{Summary of Section 3}
\begin{tcolorbox}[colback=gray!20, colframe=gray!50!black]
{\bf Types of uncertainty}: Aleatoric, epistemic \\

{\bf Representing uncertainty}: Probabilities, sets/intervals, polytopes, functions \\

{\bf Assessing uncertainty}: Variance, entropy, NLL, Mahalanobis distance, total variation distance, Hellinger distance, $\alpha$-divergences, Wasserstein distance  \\

{\bf Uncertainty quantification methods}: Ensembles, MC dropout, Laplace approximation, variational inference, MCMC,  conformal prediction, prior/posterior networks, epistemic neural networks \\

{\bf Assessing uncertainty calibration}: Entropy, NLL, Brier score, confidence plots, ECE, ACE \\

{\bf Uncertainty calibration methods}: Temperature scaling, histogram bining, Platt scaling, isotonic regression, Beta calibration, BBQ \\
\end{tcolorbox}
\end{figure}

\noindent {\bf $f$-divergences}: In probability theory, the \emph{difference between two probability distributions} can be measured by $f$-divergences. Special cases of these divergences include total variation distance, Hellinger distance, and $\alpha$-divergences. The total variation distance is simply the half the absolute area difference between the two probability density or mass functions. The Hellinger distance for discrete probability distributions $p=\{p_1, p_2, ..., p_N \}$ and $q=\{ q_1, q_2, ..., q_N \}$ is defined as,
\begin{equation}
     \frac{1}{\sqrt{2}} \sqrt{\sum_{n=1}^N(\sqrt{p_n}-\sqrt{q_n})^2}.
\end{equation}
Squaring this metric and adding $1$ gives the Bhattacharyya distance, $\sum_{n=1}^N \sqrt{p_n q_n}$. $\alpha$-divergences play a significant role in many uncertainty quantification techniques such as variational inference. A special case is KL divergence, defined as,
\begin{equation}
    \mathbb{KL}[p||q] = \sum_{n=1}^N p_n \log \left( \frac{p_n}{q_n} \right)
\end{equation}
Because the smaller the divergence is, the similar the distributions are, as a non-negative metric, $\mathbb{KL}[p\|q]=0$ divergence indicates perfect similarity between the two distributions. Note that $\mathbb{KL}[p\|q] \neq \mathbb{KL}[q\|p]$.\\

\noindent {\bf Wasserstein Distance}: In the theory of optimal transport (OT), the Wasserstein distance measures the \emph{difference between two probability distributions}. Defined by the Monge-Kantorovich theorem~\cite{villani2009optimal}. Wasserstein distance takes into account the geometry of the space.\footnote{Wasserstein is proposed in Differential Geometry whereas KL divergence is proposed in Information Theory.} Wasserstein distance can be defined for any metric space. 1-Wasserstein distance, also known as the Earth mover's distance (EMD), and 2-Wasserstein are more popular. For points $x_i$ in $p$ and $x_j$ in $q$, the latter is defined as,
\begin{subequations}
\begin{equation}
    W_2(p,q) = \min_M \sum_{i,j} M_{i,j} \| x_i - x_j\|_2^2 \label{eq:objective}
\end{equation}

subject to the constraints,
\begin{align}
    M.\mathbbm{1} &= p, \label{eq:constraint1}\\
    M^\top.\mathbbm{1} &= q, \label{eq:constraint2}\\
    M &\geq 0, \label{eq:constraintm}
\end{align}
\end{subequations}

\noindent for a coupling matrix $M$. Constrains ensure that the probability of a sample in $p$ match with all other points in $q$. Intuitively, the OT problem determines the optimal way to move the probability distribution $p$ to another $q$.

When the distributions are Gaussian or uniform distributions, the 2-Wasserstein distance has a closed form solution.\footnote{OT algorithms are implemented in POT: Python Optimal Transport library~\cite{flamary2021pot}.} 1-Wasserstein distance also has a closed-form solution for one dimensional distributions. Whenever such simplifications are not available for the integer program that is expensive to solve, the Sinkhorn algorithm can be used~\cite{cuturi2013lightspeed} by introducing an entropic regularization\footnote{$\lambda$ can be set to a large number depending on the machine precision of the computer. The larger it is, the more accurate it would be but slower the convergence is and more prone to computation errors.} term,
\begin{equation}
    \sum_{i,j} M_{i,j} \| x_i - x_j\|_2^2 - \lambda^{-1} \underbrace{ \left(- \sum_{i,j} M_{i,j} \log M_{i,j} \right)}_\text{entropy}.
\label{eq:sink}
\end{equation}

\subsection{Quantifying Uncertainty: Techniques}
\label{sec:uncquant}


Our objective is to quantify uncertainty arising from various sources discussed in Section~\ref{sec:sources}. Rather than representing the predictions as a single output, we want diverse outputs to represent the uncertainty. This predictive uncertainty can be represented probabilistically or by other means.\footnote{Historically, from the days of Jacob Bernoulli and Thomas Bayes in 1700s, uncertainty estimations were treated in a more Bayesian sense. While the efforts by Pierre-Simon Laplace and Adolphe Quetelet by applying probability in physics and social sciences in 1800s helped to disseminate statistics in various fields and Andrey Kolmogorov's work helped with mathematical formalization of these ideas~\cite{kolmogorov2018foundations}. With the successful applications of probability in Biometrics and the establishment of journals such as Biometrika, frequentists statistics became more popular in early 1900s. Since many methods were inefficient, Arthur P. Dempster and others were focusing on developing alternative techniques~\cite{walley1991statistical} to represent believes and uncertainty. With the advancement of computers in 1980s, modern Bayesian inference techniques such as MCMC were developed. They became popular especially after the application of Gibbs sampling in image processing~\cite{geman1984stochastic} and they were the most successful way to train neural networks back then~\cite{neal2012bayesian}. While MCMC techniques continued to grow, in late 1990s and 2000s Michael I. Jordan and others popularized variational inference techniques as a tractable way to estimate Bayesian posterior probabilities. With the advancement of large neural networks, Bayesian inference techniques have also been struggling. By the time of writing this article, having foundations in probability theory, conformal predictions are gaining their popularity as they are straight forward to use in deep neural networks~\cite{angelopoulos2021gentle} though they do not provide precise probabilities.} 

Central to most probability-based uncertainty techniques is the Bayes' theorem given by,
\begin{equation}
       \underbrace{q({z})}_{\text{approximate}\atop \text{posterior}}\approx \underbrace{p({z}|{y})}_{\text{posterior}}
        = \frac{ \overbrace{p({y},{z})}^{\text{joint dist.}} }{ \underbrace{p({y})}_{ \text{evidence}} } 
        = \frac{ \overbrace{({y}|{z})}^{\text{likelihood}} \overbrace{p({z})}^{\text{prior}} }
               { \underbrace{\int p({y|z})p({z}) \mathrm{d} {z} }_{\text{evidence}} }. 
    \label{eq:bayez}
\end{equation}

The prior distribution indicates our prior belief. For instance, if a robot tries to estimate where it is in a room, it can have a rough estimate, say, represented as a Gaussian distribution, of its location first. The likelihood is how we attempt to represent our data, i.e., this is data $y$, given the location $z$. By multiplying with the likelihood with prior, it accounts for the all potential locations and provides the location estimate given data, $p(z|y)$. To make the posterior a probability distribution, we have to divide this product with the evidence, also known as, marginal likelihood. 

In general, the prior can come from domain knowledge, rough estimates, or previous estimates. The latter is specially common in many sequential or iterative robotics estimation problems where the posterior at time $t$ can be used as the prior in the next time step $t+1$. If we do not know anything about the prior, we can chose a non-informative prior. This is typically an uniform distribution, or more commonly in practice, it is an exponential distribution such as a Gaussian distribution with a large standard deviation. If we do not know about the prior, we can also introduce a probability distribution over the parameters of the prior which is called a hyper-prior $p(s)$, resulting in $p(y|z)p(z|s)p(s)$. The integral in the denominator of the Bayes' theorem makes posterior estimates intractable for most choices of the prior and likelihood. The class of priors known as conjugate priors make the posterior estimation problem simpler. The choice of prior is sometimes an art; it should be good enough to to make the posterior tractable but also good enough to represent the reality. 

\begin{figure}[t]
\includegraphics[width=1.0\columnwidth]{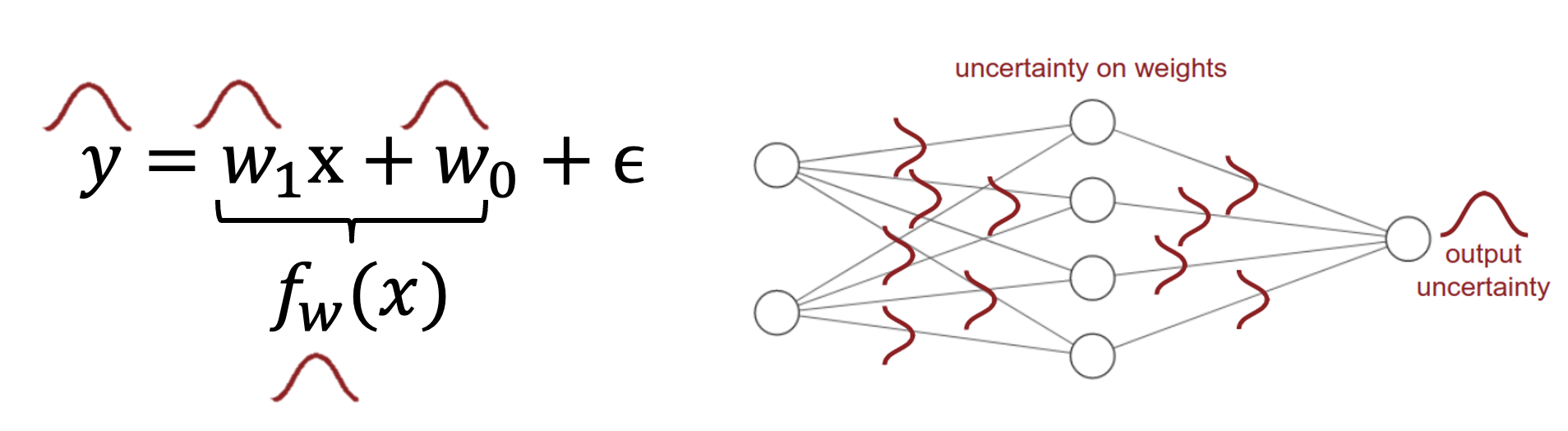}
\caption{We need many models to quantify the epistemic uncertainty. This can be done explicitly by having multiple models as in ensembles or implicitly by introducing probability distributions over the parameters (weights of the neural network) of the ML model. }
\label{fig:neural_unc}
\end{figure}

As shown in Figure~\ref{fig:neural_unc}, in most machine learning models, we introduce a prior distribution over the parameters of the model, $p(w)$, and estimate the posterior parameters given data, $p(w|\mathcal{D})$, using Bayes' theorem. Once we know this posterior distribution, the \emph{predictive probability distribution} for a new input $y_*$ can be obtained by integrating over posterior parameter estimates,
\begin{equation}
    p(y_*|\mathcal{D}) = \int \underbrace{p(y_*|w)}_\text{likelih.} \underbrace{p(w|\mathcal{D})}_\text{poster.} \mathrm{d}w
\end{equation}

This predictive distribution can also be imperially approximated by sampling from the posterior distribution, $\{ w_k \sim p(w|\mathcal{D})\}_{k=1}^K$, and computing the likelihood values, $\{p(y_*|w_k)\}_{k=1}^K$. The mean and variance can be computed from this set of predictive likelihood values.

Since posterior estimation becomes quickly intractable for complex models, mainly because of the integral in eq.~(\ref{eq:bayez}), we often have to resort to approximation techniques. Some of these approximation techniques such as variational inference and MCMC are explicit and derived from the first principle whereas some others are somewhat implicit such as MC dropout and ensembles. In what follows, we present a number of popular probabilistic and non-probabilistic techniques to quantify uncertainty in ML models. \\

\begin{figure}[]
\centering
\includegraphics[width=0.8\columnwidth]{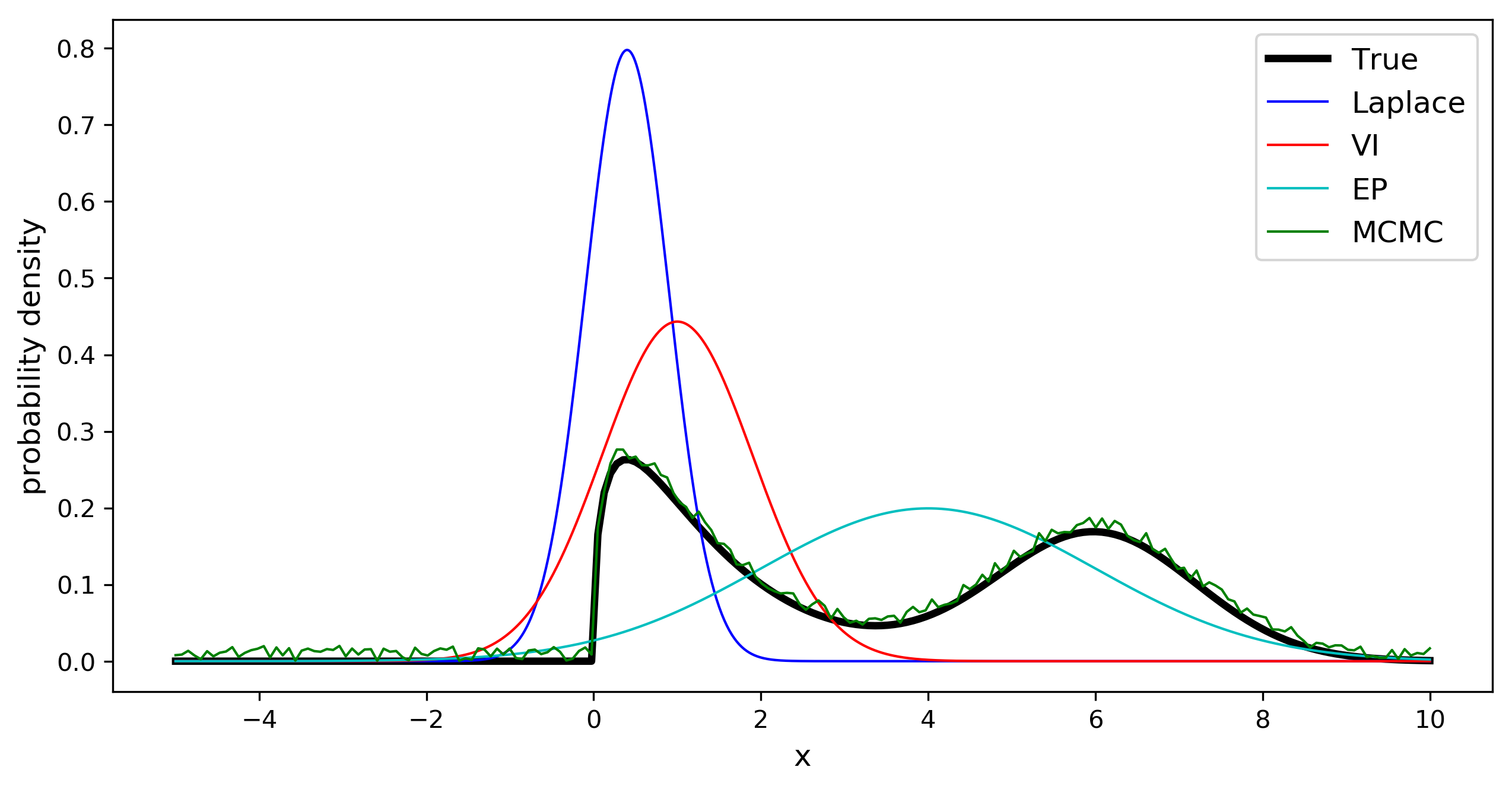}
\caption{Approximate Bayesian inference techniques. The Laplace approximation aligns with the peak. Variational inference (VI) typically aligns with one of the modes of the distribution. In contrast, expectation propagation (EP) considers the average over all modes, resulting in a more dispersed distribution. Markov Chain Monte Carlo (MCMC) can theoretically obtain the exact distribution if finitely many samples are taken.}
\label{fig2:posterior_shapes}
\end{figure}

\noindent \textbf{Ensembles}: A collection of NNs can be independently trained, ideally in parallel for computational efficiency, with different weight initializations or other changes. During test time, the mean and variance can be calculated using the outputs of different NNs. If ensemble was performed on NNs that are adversarially trained on common NN loss functions,\footnote{To be precise, the loss should be a \emph{proper scoring rule}.} then the predictive uncertainties are more accurate~\cite{lakshminarayanan2017simple,jain2020maximizing}.\\

\noindent \textbf{Monte Carlo Dropout}: To calculate uncertainty from MC dropout, the same input is passed through the network multiple times but a certain percentage of neurons are randomly disabled every time.\footnote{Dropout and MC dropout have two different objectives and work differently. The former runs in training time and works as a regularizer. The latter runs at inference time and acts as an uncertainty estimator.} Since this process implicitly induces a distribution over neural network weights, the multiple outputs represents epistemic uncertainty. This process can also be thought as variational inference~\cite{gal2016dropout}, however with a rather weak approximate posterior~\cite{folgoc2021mc}. \\

\noindent \textbf{Laplace Approximation (LA)}: The posterior is approximated with a unimodal distribution, typically a Gaussian. By using the second order Taylor approximation of the \emph{maximum a posteriori} (MAP) estimate, it places the Gaussian around the mode of the true posterior. Since this is simply a mode matching technique, it is simple and faster but less less accurate. \\

\noindent \textbf{Variational Inference (VI)}: The objective of variational inference is to learn a parameterized distribution $q(w)$ that matches with the true posterior distribution $p(w|y)$ by minimizing $\mathbb{KL}[q(w)||p(w|y)]$ for model parameters $w$. Since we do not know the true posterior, we have to derive a lower bound---typically known as the Evidence Lower Bound (ELBO)---that does not depend on the true posterior. By doing so, we convert the inference problem into an optimization problem. For instance, if the parameters of the approximate posterior follows a normal distribution, $w \sim \mathcal{N}(\mu, \sigma)$, the algorithm finds the best $\mu, \sigma$ that overlaps with $p$. Typically, estimating $w$ is computationally expensive because a model has many parameters. Therefore, we factorize the approximate posterior as $q(w)=\prod_i q_i(w_i)$ and estimate each distribution separately. This is called mean-field approximation~\cite{bishop2006pattern}. \\

Even with mean-field approximation, when we have large datasets, using the entire dataset to update parameters is computationally infeasible. By using small batches of data, we can perform Stochastic Variational Inference (SVI). This becomes specially attractive for perception tasks in robotics where the offline datasets are large or we obtain data online in batches~\cite{senanayake2017learning, senanayake2017bayesian}. All these methods require deriving the ELBO which can be mathematically taxing. Black-box variational inference (BBVI) alleviates this. Rather than obtaining an analytical form it uses automatic differentiation.\footnote{Read more about Probabilistic Programming Languages (PPL) such as Pyro, Edward (now part of TensorFlow Probability), Stan, WebPPL, and Turing.jl.} This allows learning complex distributions required for real-world robotics tasks~\cite{senanayake2018automorphing}. We also often see amortized variational inference in variational autoencoders (VAEs) in which a neural network is used to estimate the parameters of the variational distribution. However, the objective in VAEs is not obtaining outputs with uncertainty. As a separate note, when the $p$ and $q$ terms of the divergence term of VI are swapped,\footnote{Note that KL is not a symmetric metric.} $\mathbb{KL}[p(w|y)||q(w)]$, we call this Expectation Propagation (EP).\\

\noindent \textbf{Markov Chain Monte Carlo (MCMC)}: In VI, we assume a particular distribution for the posterior. Although this assumption makes computations tractable and faster, it also  limits the representation power of the posterior distribution. For instance, if a unimodal Gaussian is used for $q$, then it will not appropriately cover the true bimodal distribution $p$. Instead of assuming a parametric distribution, we can represent a distribution using samples\footnote{An infinite amount of samples represent the true distribution.}. Stein variational inference maintains a set of particles to represent the distribution---thus a nonparametric technique---and perform variational inference using them~\cite{liu2016stein}.  \\

In Monte Carlo (MC) techniques, we often assume a proposal distribution $q$ that we know of and can query. The proposal is used merely to guide sampling and needs not to match with the true distribution, although it should, ideally, encapsulate the true distribution. In rejection sampling, one of the MC techniques, we evaluate the probability density of proposal distribution $q$ and true distribution $p$ at a sampled position $i$. If $k\times q[i] < p[i]$ for some $k$, then we reject the $i^\text{th}$ sample, otherwise keep it. The collection of kept samples represent a distribution. In rejection sampling, samples are drawn independently. This uninformed sampling technique quickly becomes inefficient as it results in a lot of samples being rejected, especially when we try to estimate multimodal or high-dimensional distributions. Therefore, in techniques such as Metropolis or Metropolis-Hastings~\cite{hastings1970monte}, we sample in such a way that the next sampling step depends on the previous sample---from a Markov chain. Gibbs sampling~\cite{geman1984stochastic} samples from one or a few dimensions at a time, making it an efficient variation on Metropolis-Hasting (MH) algorithm for high-dimensional distributions. Hamiltonian Monte Carlo (HMC) allows larger jumps in MH, which results in a few samples to represent a complex distribution. No-U-Turn Sampler (NUTS) automatically fine-tunes hyperparameters in HMC~\cite{carpenter2017stan}.\footnote{These techniques are implemented in Pyro and Stan probabilistic programming language packages.} 

Langevin Monte Carlo (LMC) such as Metropolis-adjusted Langevin Algorithm (MALA)~\cite{roberts1996exponential}, takes only a single leapfrog step in HMC. Therefore, although HMC is more suitable for high-dimensional spaces, LMC is the simpler to implement and easier to use. Since traditional MCMC algorithms fall short when it comes to large datasets as the entire dataset is used to update, in Stochastic Gradient Langevin Dynamics (SGLD), parameters are updated using SGD with minibatches~\cite{welling2011bayesian}.\\

\noindent \textbf{Conformal Prediction (CP)}: Similar to MC dropout, conformal prediction is a post-hoc uncertainty estimation technique as it estimates the uncertainty of a pre-trained model using a calibration dataset and a quantile value. Conformal prediction aims at representing uncertainty with sets (or intervals). For image classification, it predicts possible set of classes (e.g., \{cat, tiger\}). The more complicated the image is or the poorer the model is, it adds more element into the set as any outcome becomes possible (e.g., \{cat, tiger, jaguar, lion\}). In regression tasks, the uncertainty is represented as an interval. Given the simplicity, conformal predictions are also well-suited for generative models such as large-language models~\cite{mohri2024language}. They are model-agnostic and provide theoretical guarantees. They have many applications in robotics as well~\cite{sun2024conformal, ren2023robots, sun2024conformal, wang2024safe, waczak2024characterizing}. CP, in its vanilla form, assumes~\emph{exchangeability}~\cite{shafer2008tutorial}, making it less straightforward to apply in certain time-dependent tasks such as robot control, though new treatments exists~\cite{barber2023conformal}. ~\citet{angelopoulos2021gentle} provide an excellent introduction for practitioners. \\

\noindent \textbf{Direct uncertainty estimation}: Techniques such as Prior Networks (PNs)~\cite{malinin2018predictive} and Posterior Networks (PostNets)~\cite{charpentier2020posterior} try to estimate the probability estimates directly while trying to maintain the properties of a predictive distribution (i.e., higher uncertainty away from data). Especially, since the PostNets use normalizing flow density estimators, they can represent complex probability distributions. These techniques are inspired from the ideas of evidential deep learning~\cite{NEURIPS2018_a981f2b7, amini2020deep}\footnote{They are based on Dempster–Shafer Theory (DST).} and have been used in applications such as autonomous driving~\cite{Itkina2022}. As another line of work, Epistemic Neural Networks (ENNs), presented as a generalized representation of other epistemic models such as Bayesian neural networks and ensembles, models the epistemic uncertainly using a small additional network called Epinet~\cite{osband2024epistemic}. \\

\noindent \textbf{Other methods and representations}: There are many other work on using alternative techniques for representing and estimating uncertainty and diversity. In work such as deep kernel learning (DKL)~\cite{wilson2016deep}, Gaussian processes are used to estimate the uncertainty. However, rather than using a standard kernel, a deep neural network is used to capture nonlinear patterns in data. Kernel learning techniques have proven to be useful in many domains~\cite{wilson2015kernel, tompkins2019black, hsu2019bayesian} and they are useful in robotics problems where using large neural networks is prohibitive. As another popular technique, particle filtering, also known as sequential Monte Carlo (SMC), is one of the most commonly used methods that helps with maintaining diverse solutions in robotics~\cite{thrun2002probabilistic}. All these are standard Bayesian techniques that were not discussed under previous methods. In addition to them, various Covariance Matrix Adaptation (CMA) techniques have been used in robot control to maintain diverse solutions~\cite{tjanaka2023pyribs,asmar2023model}. Belief functions~\cite{shafer1990perspectives} in DST, probability box (p-box)~\cite{ferson1996whereof,faes2021engineering} in engineering analysis, fuzzy sets~\cite{zadeh1965fuzzy,aditya2023pyreason} all deal with alternative uncertainty specifications and have applications to robotics~\cite{Itkina2022,tanaka2004fuzzy,bandara2016fuzzy,faes2021engineering}.

\subsection{Calibration: Are Our Uncertainties Correct?}
How do we know if the probabilities estimated by various techniques discussed in Section~\ref{sec:uncquant} are indeed accurate? Incorrect probabilities can lead to incorrect believes about safety. While various visualizations and metrics (see Section~\ref{sec:metrics}) such as accuracy vs. confidence plots, NLL, entropy, average discrepancy between class probability and ground truth (i.e., Brier score), etc. can be used to assess the correctness of uncertainty~\cite{ovadia2019can}.

Considering the robustness and flexibility, Expected Calibration Error (ECE) \cite{guo2017calibration} can be considered as the most popular method to assess calibration. In ECE, $N$ predictions are arranged into to $M$ bins each with size $1/M$ according to predicted label confidences. If average accuracy and average confidence within each bin is similar, then the model is well-calibrated. This notion can be formulated as,
\begin{equation}
    \text{ECE} = \sum_{m=1}^M \frac{|B_m|}{N} \left|
    \underbrace{\text{acc}(B_m)}_{\frac{1}{B_m}\sum_{i \in B_m} \mathbbm{1}(\hat{y}_i = y_i)} - 
    \underbrace{\text{conf}(B_m)}_{\frac{1}{B_m}\sum_{i \in B_m} \hat{p}_i} \right|
\end{equation}
\noindent where $y_i, \hat{y}_i$ and $\hat{p}_i$ are the ground truth label, predicted label, and predicted label confidence, respectively, for sample $i$ in bin $B_m$. The lower the ECE, the better calibrated the model is. Adaptive Calibration Error (ACE)~\cite{nixon2019measuring} has been proposed as an improvement to ECE.~\citet{ovadia2019can} conducted a study on the quality of uncertainty on out-of-distribution samples for various epistemic uncertainty estimation techniques. They concluded that the quality of uncertainty lowers with data shift and deep ensembles and stochastic variational inference techniques are more promising.

If a model is miss-calibrated, it needs to be calibrated. Temperature scaling and histogram binning are simple post-hoc calibration techniques. The former is parametric and the latter is nonparametric. Temperature scaling of probabilities, $\text{softmax}(\text{logits}/T)$, has a single parameter $T$ that can be tuned to minimize the ECE based on a validation dataset. A more generalized version of temperature scaling is Platt scaling, where a logistic regression model is fitted~\cite{platt1999probabilistic}. Isotonic regression involves fitting a piece-wise-constant nonparametric model. Bayesian Binning into Quantiles (BBQ)~\cite{naeini2015obtaining} is another calibration technique.\footnote{net:cal - Uncertainty Calibration Python library implements a number of these techniques.} 

\section{How Do We Leverage Uncertainty?}

Traditionally, robotics systems are designed to be modular for reasons such as interpretability and ease of designing and debugging. Nowadays, especially in research settings, the robots are also trained in an end-to-end fashion. Either way, our objective is to control the robot to achieve our objective\footnote{Our objective itself can also be uncertain.} by observing limited about of uncertain data.

\subsection{Uncertainty in Perceiving and Representing the World}
\label{sec:un_percept}

\begin{figure}[]
\centering
\includegraphics[width=0.7\columnwidth]{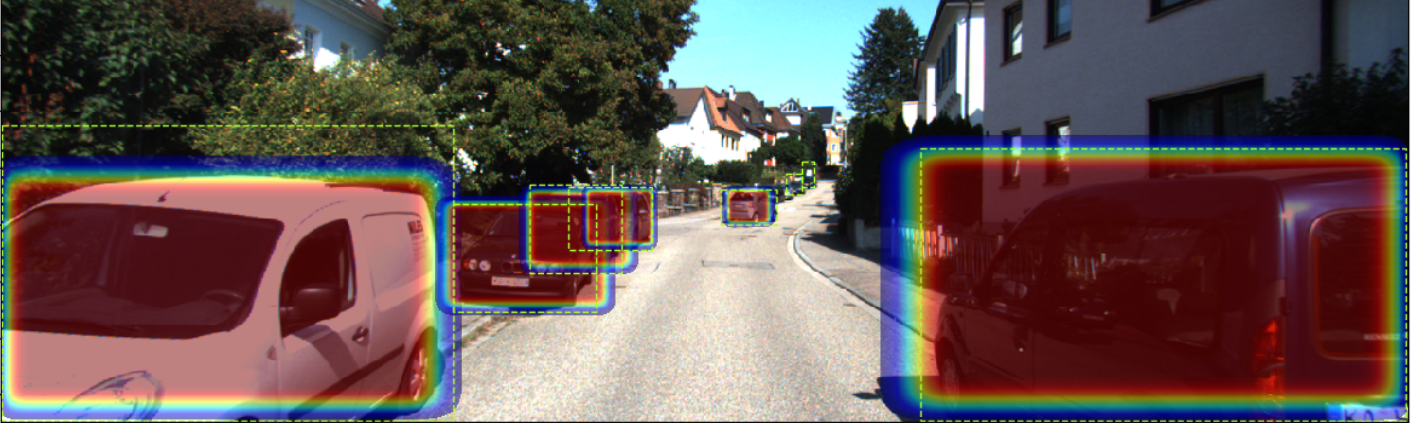}
\caption{Uncertainty in object detection in autonomous driving. Uncertainty can be associated with the detected object class or bounding boxes.}
\label{fig:class}
\end{figure}

In order for the robot to get an idea about the world, it performs various tasks.  \\

\noindent \textbf{Mapping the environment}: One of the most basic requirement for a robot to act in its environment is a map. Occupancy grid maps are one of the commonly used metric mapping technique in mobile robotics~\cite{elfes1989using,tian2023occ3d}. Considering their assumptions about discretization and inability to quantify epistemic uncertainty due to occlusions,  Gaussian process-based occupancy mapping techniques were introduced~\cite{o2012gaussian,senanayake2017bayesian}. However, considering their prohibitive computational complexity on robotic hardware, especially when the robot gathers more and more data, Bayesian Hilbert mapping (BHM)---a scalable technique that uses variational inference to estimate a distribution of maps---has been proposed~\cite{senanayake2017learning}. These techniques can estimate both aleatoric and epistemic and  uncertainty (e.g., uncertainty due to lack of data or unseen areas). There are also techniques to only estimate the aleotoric uncertainty (e.g., sensor noise) by maximizing the maximum likelihood of a Bernoulli distribution~\cite{ramos2016hilbert} or Gaussian mixture~\cite{goel2023incremental}. When the occupancy is represented as a deep neural network, DST has been used to quantify uncertainty~\cite{Itkina2022,toyungyernsub2021double}. More recently, variational inference~\cite{shen2021stochastic} and post-hoc Laplace approximations~\cite{goli2023bayes} have been used to represent uncertainty in Neural radiance fields (NeRF). A single layer NeRF functions as a BHM with random Fourier features. There have been applications of environment uncertainty quantification techniques for modeling table top objects for manipulation~\cite{wright2024v}, tactile localization and mapping~\cite{jia2022autonomous}, mapping dynamic environments~\cite{toyungyernsub2021double,min2021kernel,itkina2019dynamic,guizilini2019dynamic}, etc. \\

\noindent \textbf{Localization and tracking}: The exact location of a robot or an external agent with respect to an origin in the environment it is in is difficult to determine because of sensor limitations (e.g., low GPS signal levels), limited landmarks, occlusions, dynamics, etc. Particle filters, a sequential Monte Carlo (SMC) estimation technique, and various Kalman filters such as Extended Kalman filters (EKF), Unscented Kalman filters (UKF) have been widely used to estimate the location of the robot~\cite{durrant2001introduction,thrun2002probabilistic}. They have also been used for object tracking---determining the position over time. All these techniques focus only on aleatoric uncertainty. Also, in addition to these MC technique, variational inference has been used for localization~\cite{mirchev2018approximate, mirchev2020variational}. There are also tracking techniques that use uncertainty in deep neural networks ~\cite{zhang2022uast, zhou2021model}.  \\

\noindent \textbf{Object detection and pose estimation}: An embodied agent typically needs to identify various objects around it to act intelligently. An object in a scene is represented by a bounding box---origin, width and height---and the associated object class. Therefore, quantifying uncertainty of object detection requires, computing the uncertainty of the bounding box parameters and assigned class, for instance using MC dropout~\cite{stoycheva2021uncertainty}.~\citet{feng2021review} survey various probabilistic object detection techniques. More recently, conformal prediction has been used as a post-hoc uncertainty estimation technique for object detection~\cite{de2022object}. For most robotics tasks, merely detecting objects is not sufficient, the pose of the object needs to be estimated~\cite{xiang2017posecnn}. Estimating the poses typically requires estimating the $(x,y,z)$ position and (roll, pitch, yaw) orientation, making it an estimation problem in $\mathbb{R}^6$. Ensemble techniques~\cite{wursthorn2024uncertainty} as well as directional statistics have been used in pose estimation. In particular, probability distributions such as the von Mises-Fisher~\cite{prokudin2018deep, zhi2019spatiotemporal} and Bingham~\cite{okorn2020learning,gilitschenski2019deep} can represent uncertainty.\\

\noindent \textbf{Semantic segmentation}: Semantic segmentation is often useful in robotics tasks for delineating the foreground from background, determining the road for driving AVs, etc. Semantic uncertainty can be related to the entire class (e.g., the entire sky is identified as a river) or the class boundaries (e.g., a few pixels in the border between road and trees are incorrectly assigned). Especially, the latter case is unavoidable. Most loss functions and evaluation metrics used in semantic segmentation focus on the pixel-level semantic assignment of the entire image rather than the semantic boundaries. However, for safety critical tasks such as AVs, knowing the exact road boundaries is important. There is very high epistemic and aleatoric uncertainty in these boundaries due to ML model limitations, occlusions, etc. Similar to the mapping techniques discussed earlier in this subsection, semantic uncertainty has been quantified using relevance vector machines~\cite{gan2017sparse}, MC dropout~\cite{mukhoti2018evaluating}, etc.\\

\noindent \textbf{Imagination and future prediction}: The human ability to imagine helps us understand potential risks and efficient decisions. The action of ``pretend play'' in toddlers is considered to help with cognition, language, social, and emotional development~\cite{hashmi2020exploring}. Similarly, training robots in diverse simulations and through domain randomization help learn robust models~\cite{ramos2019bayessim}. The progress in generative AI has greatly simplified and improved the process of imagination. One special case of imagination is predicting future outcomes, which indeed is highly uncertain. Similar to humans' tendency to anticipate the future before making-decisions, robots can predict the future state of any of the tasks discussed thus far. As examples, Gaussian processes~\cite{senanayake2016spatio}, filtering~\cite{guizilini2019dynamic,itkina2019dynamic}, and convolutional LSTMs~\cite{toyungyernsub2021double,mann2022predicting} have been used to predict how the occupancy changes in space and time for the next few time steps. The somewhat similar problem of video prediction is also useful in robotics, for instance in manipulation~\cite{finn2016unsupervised,jayaraman2018time}. Most state estimation and object tracking techniques can naively be extended for predicting the future location. There are also various trajectory prediction algorithms which have proven to be useful, for instance, in autonomous driving~\cite{salzmann2020trajectron++,alahi2016social}. \\

\noindent \textbf{Agent modeling and human-AI alignment}: When robots interact with human or other robots, it is important for them to decipher the intents for successful coexistence. As a simple example, in social navigation or autonomous driving, the robot needs to guess if a human it observes would cross its path~\cite{luo2018porca}. Physical interactions~\cite{xu2019learning, dragan2013legibility}, facial expressions, and emotions can also have uncertainty. When language is used to communicate, there can be uncertainties due to delay, lack of clarity, insufficient amount of information contained in the message, etc. Uncertainty exists not just in factual information but also in other aspects of language such as prosody. Rhetorical flourish devised into modern LLMs such as GPT3.5\footnote{We believe overconfident and persuasive answers are not an inherent limitation of LLMs; rather, it should be how the agent was trained using reinforcement learning with human feedback (RLHF).} tend to provide overconfident answers. If embodied AI agents are using outputs from LLMs, their inability to express uncertainty should be taken into account. In some cases, we try to learn the various aspects of the agent such as preferences from human demonstrations, typically as as a reward function~\cite{biyik2020active,shin2023benchmarks}. There are also attempts to learn the distribution over rewards in Bayesian inverse reinforcement learning~\cite{mandyam2023kernel,ramachandran2007bayesian}. A unifying framework for imitation learning paradigms can be found in~\cite{bhattacharyya2022modeling}.  \\

\noindent \textbf{Detecting out-of-distribution (OOD) samples}: Uncertainty also helps with assessing for which inputs the model might under-perform. If the predictive epistemic uncertainty is high for a particular input, then the model is not robust in the vicinity of that input. Uncertainty has been used for OOD detection in autonomous driving~\cite{nitsch2021out}, mobile robotics~\cite{yuhas2022demo}, spacecraft pose estimation~\cite{foutter2023self}, manipulation~\cite{pmlr-v164-farid22a}, etc. Since the environments that robots operate are always subject to change, uncertainty is crucial for domain generalization and adaptation.

\subsection{Uncertainty in Planning and Control}

Our objective is to see how decision-making is affected by various sources of uncertainty. \\

\noindent \textbf{Uncertainty propagation from perception into decision-making}: Uncertainty of the world can be modeled as discussed in Section~\ref{sec:un_percept}. If a robotic system has distinct modules for perception and decision-making, as in most classical robotic setups, then the uncertainty in perception can be propagated into the decision-making modules. Our objective is to calculate the output $y=(f_\text{perc} \circ f_\text{deci})(x)$ for an input $x$ with a perception function, $f_\text{perc}: \mathcal{X} \rightarrow \mathcal{Z}$, which is then sequentially fed into a decision function, $f_\text{deci}: \mathcal{Z} \rightarrow \mathcal{Y}$. As a simple example, consider an object detection module for the perception function and a motion planner for the decision function. If the object detector is giving multiple bounding boxes with associated uncertainties, how can the decision-making module leverage all those bounding boxes? In general, if the predictive uncertainty of perception is represented as samples, as in MCMC techniques or Ensembles, each output can be run through the decision-making module. This will result in a decision distribution---multiple decisions, each with its own probability. If uncertainty is instead represented as an interval, as in conformal prediction, the two endpoints of the interval can be run separately to understand the limits of decisions. 

When the output of the perception function is a probability distribution\footnote{Say, the output of a map is the mean and standard deviation values of a normal distribution.} $p_\text{perc}$, the input to the decision function is also a probability distribution. Hence, the output of the decision-making module can be computed as,
\begin{equation}
    y = \int_{-\infty}^{+\infty} p_\text{perc}(z) f_\text{deci}(z)  \mathrm{d} z \approx \sum_{z_n \sim p_\text{perc} } f(z_n) 
    \label{eq:perdec}
\end{equation}
\noindent where $\{ z_n \}_{n=1}^N$ are samples taken from the output probability distribution of the perception module. If the probabilities are represented as a probability density function (PDF) and the integral is tractable, then the propagation becomes easier. Otherwise, we have to resort to a numerical integration technique. By sampling from the predictive probability distribution of perception and then evaluating the decision function on those samples, as shown in eq.~(\ref{eq:perdec}), is a straightforward simplification. If the decision-making module is also setup to compute uncertainty, then $y$ is also a probability distribution.\\

\noindent \textbf{The use of uncertainty in exploration for better world representations}: In section~\ref{sec:un_percept}, we discussed the importance of building a map of the environment. The objective of uncertainty quantification for mapping is understanding which areas of the environment we do not know about. This information can be used by the robot for exploration---to gather more information about the environment. Some applications of such exploration include a robot mapping an indoor environment, environmental monitoring using a drone~\cite{marchant2012bayesian}, or subterranean or extraterrestrial navigation~\cite{kim2021plgrim,Peltzer2022arxiv}. For this purpose, probabilistic frontiers~\cite{yamauchi1997frontier}, Bayesian optimization (BO) with Gaussian processes~\cite{marchant2012bayesian}, Partially Observable Markov Decision Process (POMDP) solvers~\cite{morere2017sequential}, and reinforcement learning (RL)~\cite{garaffa2021reinforcement} have been used. Frontiers and BO are categorized under myopic exploration strategies as they are generally one-step look-ahead planners. \\

\noindent \textbf{Uncertainty in learning a policy}: Uncertainty is widely studied in planning and scheduling~\cite{jensenfault,nagami2024state} as well as robot motion planning~\cite{lai2020bayesian}. A robot takes a sequence of decisions to control itself. This process can be modeled as a Markov decision process. If observations are not fully observable, we consider partially observable decision-making processes (POMDPs). POMDPs are widely used in manipulation~\cite{curtis2023task}, autonomous driving~\cite{kruse2022uncertainty}, aerospace control~\cite{kochenderfer2015decision}, etc. ~\citet{kochenderfer2015decision,kochenderfer2022algorithms} list various solvers\footnote{Various solvers are implemented in POMDPs.jl~\cite{egorov2017pomdps}} for POMDPs or belief space planning~\cite{platt2010belief}. Note that the typical POMDPs do not take into account the epistemic uncertainty; estimating the epistemic uncertainty requires maintaining a distribution over the MDP which is intractable. 

We typically use stochastic policies as the outcomes of our actions are not certain. This randomness also help with exploring different possibilities rather than getting stuck in a local optimum. When the model of the environment dynamics is unknown or too complex to model, we have to resort to \emph{model-free reinforcement learning (RL)}. Therefore, it implicitly handles a limited aspects of uncertainty of the environment.

\begin{figure}[]
\centering
\includegraphics[width=\columnwidth]{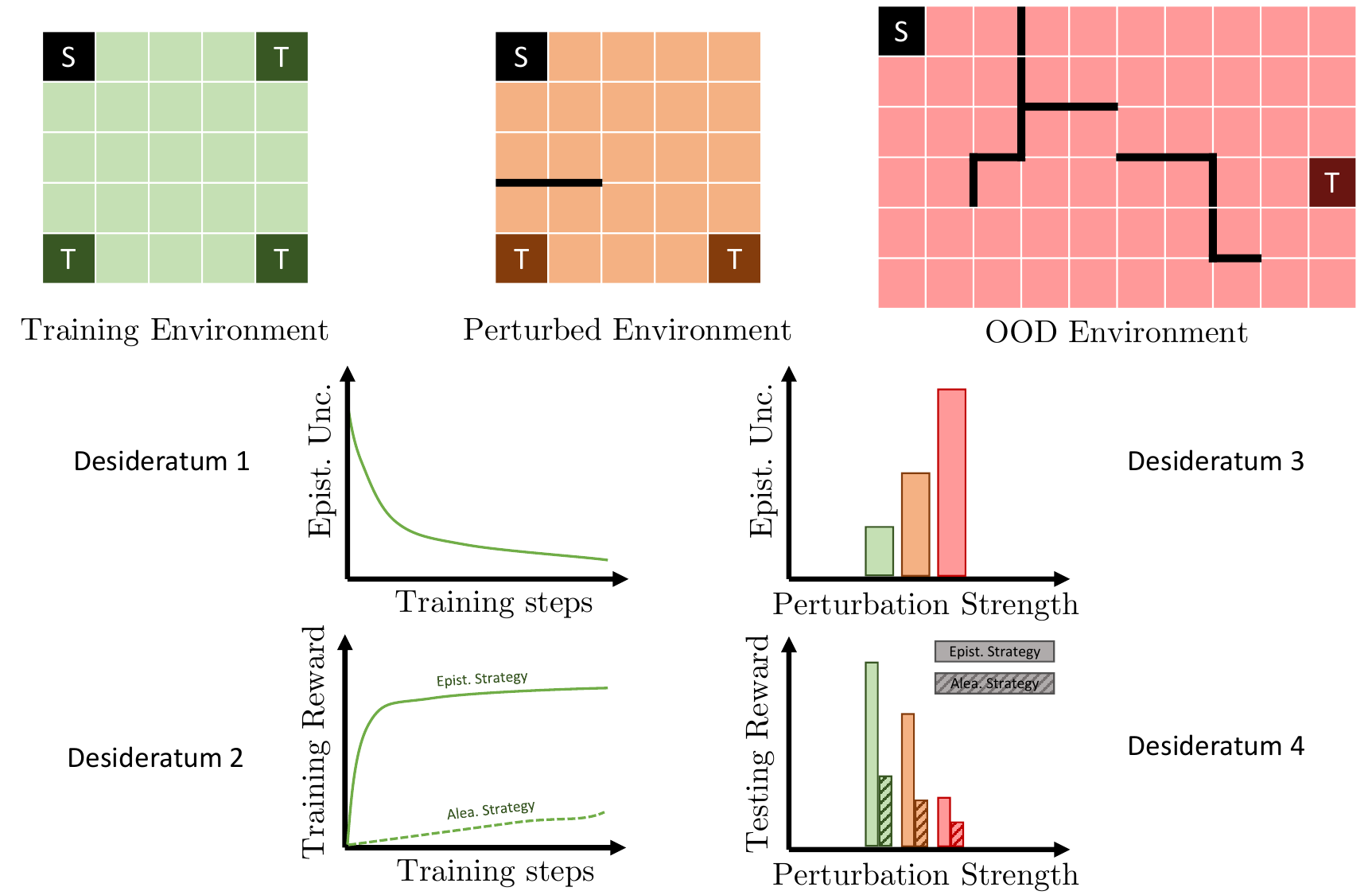}
\caption{Uncertainty in deep RL. Green, orange, and red show environments with increasing difficulty. Four desiderata for the desired behavior of uncertainty is discussed in~\cite{charpentier2022disentangling}.}
\label{fig:rl}
\end{figure}

Aleatoric and epistemic uncertainty can help improve deep RL in various ways. Following~\citet{charpentier2022disentangling}, as illustrated in Figure~\ref{fig:rl}, we discuss four desiderata for the desired behavior of uncertainty in deep RL. As the number of training steps increases, the agent's epistemic uncertainty should decrease as the agent collects more information about the environment (desideratum 1). Simultaneously, agent should collect more rewards with an epistemic strategy compared to an aleatoric strategy\footnote{Aleatoric and epistemic strategies involve sampling from aleatoric and epistemic uncertainty distributions, respectively, to select actions.} (desideratum 2). In the real-world, the environment we train our robots is not the same as we deploy. It sometimes tend to have little (e.g., orange environment in Figure~\ref{fig:rl}) or even completely different perturbations (red environment in Figure~\ref{fig:rl}). As we increase the complexity of perturbations, the epistemic uncertainty should be higher (desideratum 3) as the agent does not know about these new environments. Simultaneously, the rewards it collects should go lower (desideratum 4). A comparative analysis on the effect of deep RL algorithms is provided in \cite{charpentier2022disentangling}. The paper empirically and theoretically concluded that compared to an $\epsilon$-greedy or aleatoric strategy, sampling using an epistemic strategy receives higher rewards and maintains high generalization performance in OOD domains. This characterization of aleatoric and epistemic uncertainty in deep RL helps with generalization, sample efficiency, and knowing where the policy will not work.

Uncertainty is also studied in control theory. Bayesian nonparametric techniques such as Gaussian processes~\cite{akametalu2014reachability} and set-based techniques such as conformal prediction~\cite{lindemann2023safe} have gained popularity or control tasks because they can provide theoretical guarantees, data efficient, and simple to use. Other set-based techniques such as backward reachability has also helped with reliability and safety~\cite{bansal2017hamilton,herbert2017fastrack}. When uncertainty in parameters or unmeasured disturbances exist, Robust Model Predictive Control (Robust MPC) strategies help to maintain the stability of the system~\cite{bemporad2007robust}. Uncertainty in Robust MPC is typically handled as bounded sets such as polytopes or ellipsoids. Conditional Value at Risk (CVaR) also has been used in robot robust control~\cite{vincent2023guarantees}, trajectory optimization~\cite{lew2023risk}, and motion planning~\cite{hakobyan2019risk} to address risks and uncertainties. \\
 
\section{Conclusions}
We discussed how uncertainty and diversity are crucial in evaluative and generative paradigms of embodied AI. We highlighted that most classic robotics tasks focus primarily on aleatoric uncertainty. Although quantifying epistemic uncertainty is less straightforward, doing so can enhance robustness, data efficiency, and the capacity for imagination in embodied AI agents.

\small

\end{document}